\documentclass[11pt]{article}

\usepackage[preprint]{acl}

\usepackage{times}
\usepackage{latexsym}

\usepackage[T1]{fontenc}

\usepackage[utf8]{inputenc}

\usepackage{microtype}

\usepackage{inconsolata}

\usepackage{graphicx}

\newcommand{\eg}{\textit{e}.\textit{g}.}

\usepackage{amsmath}
\usepackage{mathtools}
\usepackage{amsthm}
\usepackage{amssymb}

\usepackage{tabularx}
\usepackage{booktabs} 
\usepackage{multirow}

\title{LinguaGame: A Linguistically Grounded Game-Theoretic Paradigm for Multi-Agent Dialogue Generation}

\author{
 \textbf{Yuxiao Ye\textsuperscript{1}},
 \textbf{Yiming Zhang\textsuperscript{2} \thanks{Equal contribution.}},
 \textbf{Yiran Ma\textsuperscript{3}\footnotemark[1]},
 \textbf{Huiyuan Xie\textsuperscript{1}},
\\
 \textbf{Huining Zhu\textsuperscript{4}},
 \textbf{Zhiyuan Liu\textsuperscript{1} \thanks{Corresponding author.}},
\\
\\
 \textsuperscript{1}Tsinghua University,
 \textsuperscript{2}University of California, Berkeley,
 \textsuperscript{3}Peking University,
\\
 \textsuperscript{4}East China University of Political Science and Law
\\
 \small{
   \textbf{Correspondence:} yeyuxiao@mail.tsinghua.edu.cn
 }
}

\begin{document}
\maketitle
\begin{abstract}
Large Language Models (LLMs) have enabled Multi-Agent Systems (MASs) where agents interact through natural language to solve complex tasks or simulate multi-party dialogues.
Recent work on LLM-based MASs has mainly focused on architecture design, such as role assignment and workflow orchestration. 
In contrast, this paper targets the interaction process itself, aiming to improve agents' communication efficiency by helping them convey their intended meaning more effectively through language.
To this end, we propose \textbf{\emph{LinguaGame}}, a linguistically-grounded game-theoretic paradigm for multi-agent dialogue generation.
Our approach models dialogue as a signalling game over communicative intents and strategies, solved with a training-free equilibrium approximation algorithm for inference-time decision adjustment.
Unlike prior game-theoretic MASs, whose game designs are often tightly coupled with task-specific objectives, our framework relies on linguistically informed reasoning with minimal task-specific coupling. 
Specifically, it treats dialogue as intentional and strategic communication, requiring agents to infer what others aim to achieve (intents) and how they pursue those goals (strategies). 
We evaluate our framework in simulated courtroom proceedings and debates, with human expert assessments showing significant gains in communication efficiency.
\end{abstract}

\section{Introduction}

A Multi-Agent System (MAS) consists of autonomous, interacting agents that learn from their environment and peers to make decisions and collaboratively solve tasks \citep{dorri2018multi}.
With the recent advancement of Large Language Models (LLMs), MASs can now be constructed by assigning each agent (instantiated as an LLM) a role and enabling them to communicate with one another in natural language \citep{park2023generative,guo2024large,han2024llm}.
LLM-based MASs have recently been applied to collaborative problem-solving (\eg \ software development and collective reasoning) and to multi-party dialogue simulation (\eg \ courtroom proceedings) \citep{cui2023chatlaw,chen2024agentcourt,Chen2024agentverse,qian-etal-2024-chatdev}.

Most existing approaches focus on tailoring MAS architectures to downstream tasks \citep{li2024survey,dang2025multi,shengbinyue2025multi}, with improvements driven largely by role-assignment prompts and workflow orchestration. 
However, such architecture-centric methods often overlook the underlying dynamics of agent interaction.
To address this limitation, we propose \emph{LinguaGame}, a game-theoretic paradigm that models multi-agent dialogue as a signalling game over communicative intents and strategies.

Game theory provides a principled framework for modelling strategic interaction among agents \citep{barron2024game}.
Existing game-theoretic MASs typically design games tightly coupled with specific tasks, or even directly instantiated as the tasks themselves, which limits their generalisability \citep{peters2024contingency,he2025generative}. 
In contrast, LinguaGame grounds game design in linguistic reasoning: agents infer each other's communicative intents and strategies.
That is, they reason about what others aim to achieve through their utterances and how language is strategically formulated to serve those goals.
This linguistically grounded formulation enables flexible game design with minimal task-specific adaptation.


Our approach is also computationally efficient at inference time for multi-agent dialogue generation.
Whereas conventional MASs improve through model retraining or repeated experience accumulation \citep{cheng2024self,chen2024agentcourt,qian2024experiential}, LinguaGame enables inference-time decision adjustment via a training-free equilibrium approximation algorithm. 
This plug-and-play mechanism reduces computational cost and deployment overhead, making it practical for diverse MAS settings.

Our evaluation spans simulated courtroom proceedings and debate scenarios.
Unlike tasks such as code completion or question answering, which primarily focus on end results, these settings emphasise open-ended dialogue, where the conversational process itself is of substantive importance.
This makes them well suited for evaluating the quality of multi-agent dialogue generation.
Human expert evaluations demonstrate that LinguaGame significantly improves dialogue quality by enhancing agents' communication efficiency.
Although our experiments focus on legal and argumentative settings, the core mechanism ``modelling dialogue as the interplay of communicative intents and strategies'' reflects a fundamental aspect of natural language use.
As such, our approach has the potential to generalise to diverse multi-agent dialogue generation scenarios.

Our contributions are as follows:
\begin{itemize} \setlength{\itemsep}{0pt}\setlength{\parskip}{0pt}
    \item We propose LinguaGame, a novel game-theoretic paradigm for multi-agent dialogue generation that models communication as a signalling game grounded in linguistic reasoning.
    \item Our framework decouples game design from task-specific objectives and relies on pragmatic aspects of communication, enabling broad generalisation across tasks with minimal adaptation.
    \item We introduce a training-free equilibrium approximation algorithm that guides agent decisions in inference time. It can be integrated in a plug-and-play manner, without requiring model retraining or iterative interaction.
    \item We demonstrate the effectiveness of our framework through experiments in simulated courtroom and debate scenarios, showing significant improvements in communication efficiency.
\end{itemize}

\section{Related Work}

We position our work at the intersection of three lines of research: LLM-based MASs in general, game-theoretic MASs in particular, and behaviour improvement mechanisms in MASs.

\paragraph{LLM-based MAS}

As an emerging research direction, most existing work on LLM-based MASs focuses on designing system structures to enable these systems to perform specific tasks.
This typically involves refining role assignment and workflow orchestration.
For instance, ChatDev \citep{qian-etal-2024-chatdev}, a software development MAS, defines five agent roles, namely CEO, CTO, reviewer, tester, and programmer, and coordinates their collaboration via an inception prompting mechanism \citep{li2023camel}.
In our work, we do not pursue performance gains through such structural refinement.
Instead, we take a closer look at the internal dynamics of MASs, focusing on agent communication.
Our aim is to improve overall dialogue generation quality by enhancing agents' communication efficiency.

\paragraph{Game-theoretic MAS}

Although MASs are inherently suited for game-like interactions involving multiple players, research on game-theoretic MASs remains limited. 
Existing work in this area typically incorporates game theory by constructing task-specific games that are closely aligned with or directly represent the target tasks themselves.
For example, in the task of robotic motion planning for autonomous driving, \citet{peters2024contingency} formulate a contingency game in which MAS agents select strategic driving trajectories based on possible future intents of others. 
For the task of mobile networking optimisation, \citet{he2025generative} design an MAS where agents play games which are subtasks of networking optimisation.
In contrast, our game formulation is grounded in general linguistic principles derived from Speech Act Theory \citep{searle1969speech}, which views utterances as expressions of speaker intent.
This abstraction enables our method to generalise more easily across tasks.

Within game-theoretic approaches to language use, our work is most closely related to the Rational Speech Act (RSA) framework \citep{frank2012predicting,goodman2016pragmatic}.
It models pragmatic communication as a process in which speakers choose utterances to convey intended meanings and listeners infer those meanings.
RSA-style models are typically instantiated in tightly constrained reference games or other toy domains with explicitly defined utterance and meaning spaces.
For example, \citet{goodman2016pragmatic} study reference games and scalar implicature tasks over small, enumerated meaning and utterance sets.
\citet{monroe2017colors} apply RSA-style reasoning to a colour reference game with discretised meanings and a restricted descriptive vocabulary.
In contrast, LinguaGame shifts pragmatic reasoning from utterance-level inference to intent-strategy abstraction, enabling RSA-style models to be applied in realistic, LLM-mediated multi-agent dialogue.

\paragraph{MAS Behaviour Improvement}

Existing LLM-based MASs often improve agent behaviour through learning-based mechanisms, such as model retraining or iterative experience accumulation.
For example, in the word-guessing MAS of \citet{cheng2024self}, agents repeatedly play the game to collect successful dialogue episodes, which are then used to fine-tune the LLM via offline reinforcement learning.
In the MAS by \citet{chen2024agentcourt}, a lawyer agent records insights from each simulated courtroom proceeding into an experience database which it consults in subsequent rounds, gradually improving argumentative performance through accumulated experience.
These approaches require either costly retraining or repeated execution of the full system.
In contrast, LinguaGame operates entirely at inference time, using a training-free equilibrium approximation to re-rank candidate utterances, without updating model parameters.


\section{Method} 

We propose LinguaGame, a linguistically-grounded game-theoretic paradigm for multi-agent dialogue generation. 
The framework operates on top of a standard MAS foundation, augmenting it with a game-theoretic formulation of agent interaction and a training-free algorithm for equilibrium approximation.

\subsection{MAS Structure}


\paragraph{MAS for Courtroom Proceedings} We configure our MAS for simulated courtroom proceedings in accordance with first-instance proceedings under Chinese civil law.
In this configuration, the procedure comprises five sequential stages that involve interactive or strategic communication: 1) Evidence Presentation and Cross-Examination, 2) Court Investigation and Questioning, 3) Court Debate, 4) Final Statement, and 5) Judgment Announcement.


This MAS retains three primary roles in courtroom proceedings: judge, plaintiff lawyer, and defendant lawyer.
Each simulated courtroom proceeding is grounded in a real-world written judgment drawn from Chinese court decisions.
This material serves as the shared factual basis and is made available to all roles throughout the proceeding.
Prior to the start, the judge is briefed on the five-stage procedure. 
The judge then initiates the proceeding and manages turn-taking by deciding which role speaks next after each contribution. 
Within each stage, the full conversation history is visible to all roles.
At the end of each stage, the entire dialogue from that stage is summarised by an LLM. 
This summary is then provided to all roles as shared context for the subsequent stage.

\paragraph{MAS for Debates}  
We configure our MAS for simulated debates as open-ended exchanges between two opposing roles on a given controversial proposition (\eg \ ``\emph{abortion should be legalised}'').  
The two roles are defined symmetrically: a proponent, who argues in favour of the proposition, and an opponent, who argues against it.
To encourage substantive discussion, each arguer is required to develop arguments from at least two distinct aspects.

Unlike courtroom proceedings, debates are not governed by a fixed procedural structure.  
Instead, the arguers alternate turns, presenting claims, counterarguments, and rebuttals in an open-ended dialogue.
A debate concludes when the dialogue converges naturally to closure, as determined by an LLM-based assessment. 
Throughout the debate, both arguers have access to the complete conversation history. 

\subsection{Game Formulation}

We adopt a game-theoretic perspective to model agent interaction as strategic communication.
Specifically, we use the \emph{signalling game} framework \citep{lewis2008convention,sobel2020signaling}.

\paragraph{Signalling Game} 
In a standard signalling game, a \emph{sender} selects a \emph{signal} to convey \emph{private information} that is visible only to them, and a \emph{receiver} interprets the signal to infer that hidden information and respond accordingly.
Each player in a signalling game is assigned a \emph{utility function}, which quantifies how well the outcome of the interaction aligns with their goals.


\begin{figure*}[t]
\centering
\includegraphics[width=\textwidth]{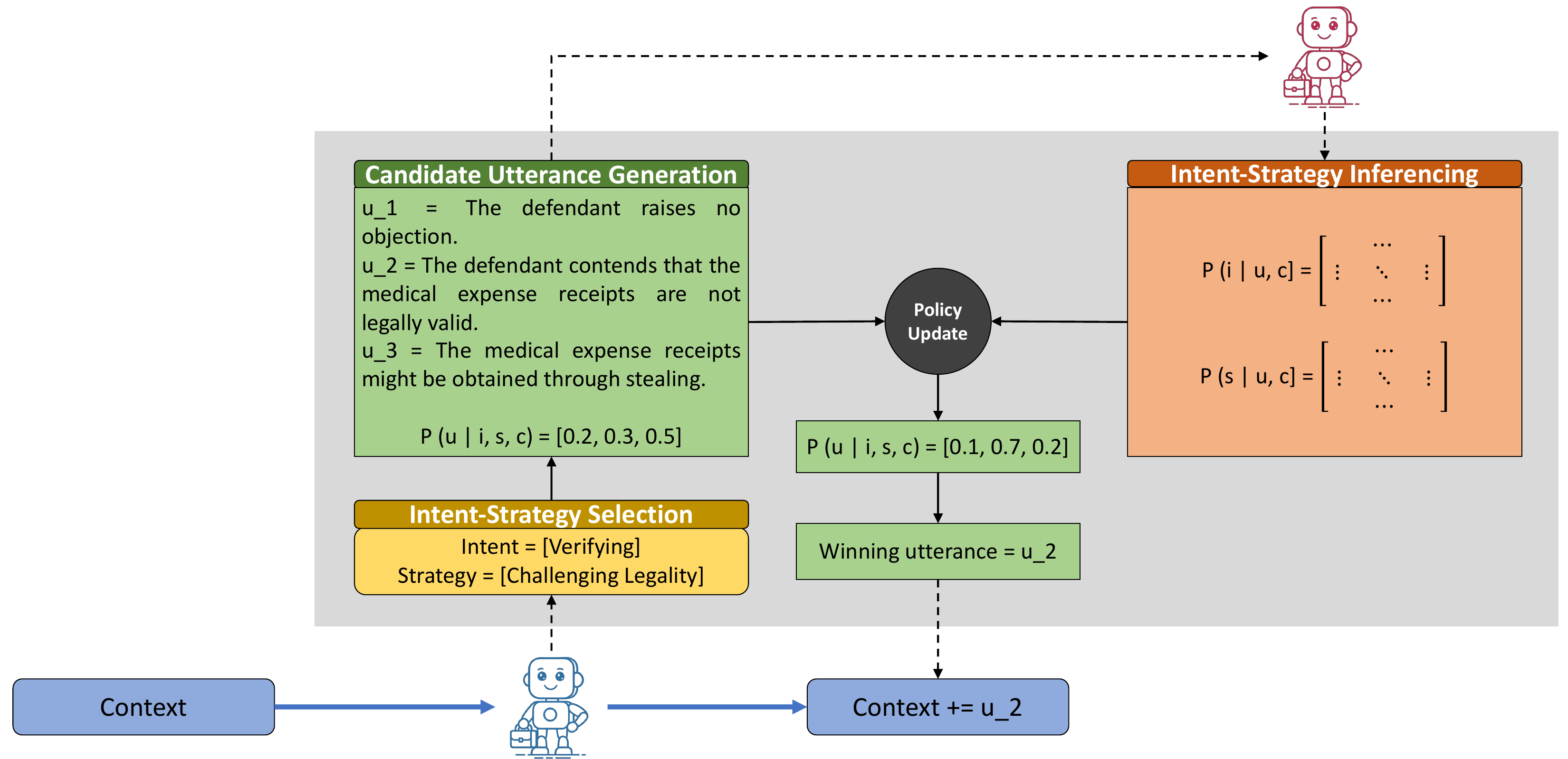}
\caption{Integration of the signalling game into the MAS in LinguaGame.
The sender (the agent in blue) generates candidate utterances, which are interpreted by the receiver (the agent in red).
The utterance with the highest probability under the updated sender policy is appended to the dialogue context.
All intermediate information produced during the gaming process (in the grey box) is discarded once the game concludes.}
\label{figs/game}
\end{figure*}  


In our adaptation of the signalling game to multi-agent conversations, a game is instantiated immediately before an agent generates its next utterance. 
The agent about to speak acts as the sender, and the agent scheduled to reply acts as the receiver. 
The signal corresponds to the utterance generated by the sender.


\paragraph{Intent-strategy as Private Information} 
To define the private information that needs to be inferred, we draw on Speech Act Theory \citep{searle1969speech}, which holds that utterances are not merely vehicles of content but also convey speakers' intentions.

Building on this idea, we model the sender's private information as a pair of communicative intent and strategy.
Intents capture what the speaker aims to achieve, while strategies capture how the intent is realised linguistically. 
For courtroom proceedings we define nine intents and twelve strategies; for debates we define six intents and eight strategies (Appendices~\ref{sec:app/int}-\ref{sec:app/str}).
We do not devise strategies for every intent, but only for those where agents can pursue their goals in substantially different ways.
Intents such as {\sc Managing Proceeding} and {\sc Ruling}, which are procedural or outcome-driven in nature, do not require further strategic differentiation. 


The design of the intent and strategy inventories is grounded in domain expertise and established argumentation theory.
For courtroom proceedings, they were derived through consultation with practising Chinese judges and lawyers, while for debates they follow established Argumentation Schemes \citep{walton2008argumentation}.
Aligning the intent and strategy inventory with the communicative conventions of specific scenarios requires only minimal effort. 
In our case, defining them for each scenario took around two hours.

\subsection{Game-MAS Integration}

Figure~\ref{figs/game} illustrates how the signalling game is integrated into the MAS. 
Given a conversational context $c$, the sender $S$ first selects an intent-strategy pair $(i, s)$. 
Conditioned on $c$ and $(i, s)$, $S$ then generates a finite set of candidate utterances $\{u_1, u_2, u_3, \dots\}$.
For each candidate utterance $u$, the receiver $R$ interprets it and independently
\footnote{Although strategies depend on intents by design, $R$ infers them independently: intents are chosen from the full space without regard to strategies, and strategies are chosen from the subset of the ground-truth intent without feeding back into intent inference. 
This preserves the hierarchical relation while avoiding the combinatorial complexity of joint inference.} 
infers the underlying intent and strategy.
Key prompts used in this process are shown in Appendix~\ref{sec:app/prompt}.

Based on the outcome of this inference, the policies of both $S$ (utterance generation) and $R$ (intent and strategy inference) are updated through the equilibrium approximation procedure.
The candidate utterance with the highest probability under the updated sender policy, given $(i, s)$, is selected as the \emph{winning utterance}.

After the game concludes, only the winning utterance is appended to the dialogue context and passed on as part of the conversation.
All intermediate information used during the game (such as candidate utterances, intents, and strategies) is discarded, ensuring that agents do not retain privileged information when continuing the dialogue.
At each dialogue turn, this process constitutes an inner-loop equilibrium search over sender and receiver policies induced by the base language model and inference prompts, whose sole output is the selected utterance appended to the dialogue.

\subsection{Equilibrium Approximation}

Signalling games are typically solved by identifying a \emph{Nash equilibrium}, a stable configuration in which no player has an incentive to unilaterally change how they convey or interpret messages.
In other words, solving the game requires finding a set of policies such that each player's choices maximise their expected utility.

\paragraph{Utility Functions} 

Although courtroom and debate settings are adversarial in terms of overall goals, from the perspective of communication efficiency the interacting roles must cooperate to ensure mutual understanding.
We therefore formulate this signalling game as a \emph{cooperative game}, in which both the sender and the receiver get a reward when the private information is correctly inferred.
Thus, both the sender and the receiver share a same part in their utility functions, defined as follows:
\begin{equation} \label{eqs/U_shared}
\begin{split}
U_{shared} = & \sum_{i,s}{p(i,s)} 
\sum_{u}{\pi_S(u \mid i,s)} 
 \bigl[
w\pi_{R_i}(i \mid u) \\
& + (1-w)\pi_{R_s}(s \mid u)
\bigr]
.
\end{split}
\end{equation}

\noindent Here $p(i,s)$ denotes the prior distribution of an intent-strategy pair $(i, s)$ given the context\footnote{For brevity, we omit explicit conditioning on the context $c$ in all equations where it is understood.}. 
Note that the prior distribution of intent-strategy pairs appears in the utility definitions only for the purpose of policy optimisation. This information is never revealed to the receiver during inference.
The sender's policy, $\pi_S(u \mid i,s)$ specifies the probability of choosing utterance $u$ conditioned on $(i,s)$ and $c$. 
The receiver's policy $\pi_{R_i}(i \mid u)$ is for inferring the intent, and $\pi_{R_s}(s \mid u)$ is for inferring the strategy.
The scalar $w \in [0,1]$ is a weighting hyperparameter that balances the relative importance of intent and strategy in the overall utility calculation.

Following \citet{jacobconsensus}, we add a KL-regularisation term into the utility function to prevent abrupt or unstable policy updates: 
\begin{equation} \label{eqs/U_S}
\begin{split}
U_S(\pi_S \mid \pi_R) 
= 
-\lambda  \mathrm{KL}\bigl(\pi_S \big\| \pi_S^{0}\bigr) 
+
U_{shared}
, 
\end{split} 
\end{equation}
\begin{equation} \label{eqs/U_R}
\begin{split}
U_R(\pi_R \mid \pi_S) 
= 
-\lambda  \mathrm{KL}\bigl(\pi_R \big\| \pi_R^{0}\bigr) 
+
U_{shared}
,
\end{split}
\end{equation}

\noindent where $\pi_S^{0}$ and $\pi_R^{0}$ denote the initial reference policies, and $\lambda > 0$ is a  hyperparameter controlling the strength of regularisation.

\paragraph{Utility Optimisation} 

In order to compute a Nash equilibrium, we optimise the utility functions with no-regret learning \citep{daskalakis2021near}, in which policies are iteratively updated and gradually converge to equilibrium.

We adopt the piKL algorithm \citep{jacob2022modeling} to implement no-regret learning, where each player iteratively updates its policy by optimising the expected utility.
Importantly, this equilibrium approximation is performed entirely as an inner-loop optimisation at inference time and does not involve additional LLM calls.
Further details can be found in Appendix~\ref{sec:app/update}.

As the update process converges, the players are regarded as reaching an approximate Nash equilibrium.
The utterance chosen for MAS generation is then the candidate with the highest probability under the sender's final policy given the ground-truth intent-strategy pair.

\section{Experiments}

Building on the courtroom proceeding scenario introduced in the Method section, we conduct experiments to evaluate the effectiveness of LinguaGame.
Specifically, we assess whether it improves communication efficiency in multi-agent dialogue generation through expert evaluation of the generated dialogues.

\subsection{Experimental Setup}

We choose Qwen2.5-32B \citep{qwen2025qwen25technicalreport} as the base LLM in the experiment. 
In our pilot studies, we experimented with several alternatives, including MiniCPM4-8B \citep{team2025minicpm4}, Qwen2.5-7B \citep{qwen2025qwen25technicalreport}, as well as Meta-Llama-3.1-8B-Instruct and Meta-Llama-3.1-70B-Instruct \citep{grattafiori2024llama}.
We observed that smaller models (7B and 8B variants) exhibited notably weaker instruction-following capabilities, while the two larger models performed comparably in instruction adherence and dialogue quality.

For the hyperparameters used in the signalling game, we set the number of candidate utterances to $3$.
Following a grid search in our pilot experiments, we set the weighting hyperparameter $w$ to $0.5$, or to $1.0$ when the selected intent lacks associated strategies.
The KL-regularisation weight $\lambda$ and the policy update learning rate are both set to $0.1$.
The equilibrium approximation runs for 5000 optimisation rounds.
While these hyperparameters could be further tuned, hyperparameter optimisation is not the focus of this paper.
The current configuration is sufficient to produce meaningful results for evaluating the framework.

For courtroom proceedings, we randomly select 50 cases from China Judgments Online \citep{china_judgments_online}, each associated with a distinct cause of action. 
From the written judgment of each case, we extract the party information\footnote{Only non-identifiable attributes such as gender and age are preserved without modification.} and a brief case description, which serve as input for the simulation.
For debates, we randomly select 50 controversial propositions from the Quora Argumentation Mining dataset \citep{ye2024computational}, spanning 20 topics such as politics, society, and technology, and use them as input for the simulation.

\subsection{Systems}

Our method is designed as a plug-in interaction layer that can be applied on top of existing MAS dialogue generation systems without modifying their underlying architectures.
Accordingly, we adopt a stepwise ``before-after'' evaluation design, comparing the same base system with progressively added components to isolate the contribution of each component.
We compare the following three systems:
\begin{itemize} \setlength{\itemsep}{0pt}\setlength{\parskip}{0pt}
    \item \textbf{\emph{SDMAS}}: The standard MAS setup described in the Method section.
    \item \textbf{\emph{ISMAS}}: The standard MAS equipped with our intent-strategy design. 
    In this setting, each agent generates utterances conditioned on a given intent-strategy pair in a Chain-of-Thought (CoT) \citep{wei2022chain} manner.
    \item \textbf{\emph{LGMAS}}: The standard MAS integrated with our full LinguaGame paradigm, which incorporates both the intent-strategy design and the game-theoretic modelling.
\end{itemize}

To evaluate the game-theoretic utterance selection in LinguaGame against a strong inference-time alternative, we introduce an LLM-based utterance re-ranking baseline.
This baseline selects the final utterance from the same set of candidate utterances generated by \emph{LGMAS}, using a single-shot judgment by an LLM.
Specifically, given the full dialogue history up to the current turn and a fixed set of candidate utterances, the LLM is prompted to select the most appropriate next utterance.
The re-ranking decision is made independently at each turn.
We experimented with Qwen2.5-32B, Meta-Llama-3.1-70B-Instruct, and DeepSeek-V3-0324 \citep{deepseekai2024deepseekv3technicalreport} for this baseline.

\begin{table*}[t]
\centering
\fontsize{10.0pt}{\baselineskip}
\selectfont
\begin{tabular}{@{}c|lrrr|rrr|rrr@{}}
\toprule[1pt]
\multicolumn{1}{c}{}                         &     & \multicolumn{3}{c}{\emph{SDMAS}}   & \multicolumn{3}{c}{\emph{ISMAS}}  & \multicolumn{3}{c}{\emph{LGMAS}}   \\ 
\cline{3-11}
\multicolumn{1}{c}{}                         &     & Court & Debate & \multicolumn{1}{r}{Overall}  & Court & Debate & \multicolumn{1}{r}{Overall}  & Court & Debate & Overall   \\ 
\midrule
\multirow{2}{*}{Statistics}                  & utt/dialogue     & 180  & 48     & 114      & 124  & 34     & 79       & 126  & 36     & 81 \\
                                             & token/utt        & 80   & 135    & 108      & 65   & 120    & 93       & 70   & 125    & 98 \\
\midrule
Linguistic                                   & Clarity          & 4.10 & 4.15   & 4.13     & 4.18 & 4.22   & 4.20     & 4.36 & 4.42   & 4.39 \\
form                                         & Conciseness      & 3.55 & 3.62   & 3.59     & 3.73 & 3.69   & 3.71     & 4.12 & 4.19   & 4.16 \\
\midrule                                            
Content                                      & Argument         & 3.18 & 3.27   & 3.23     & 3.23 & 3.34   & 3.29     & 3.83 & 3.92   & 3.88 \\
quality                                      & Tactic           & 3.28 & 3.39   & 3.34     & 3.35 & 3.42   & 3.39     & 3.96 & 4.05   & 4.01 \\
\bottomrule[1pt]
\end{tabular}
\caption{Statistics of the generated dialogues and results for the dialogue-level evaluation.}
\label{tabs/dialogue_eval}
\end{table*}

\subsection{Evaluation}

We conduct manual evaluations of the dialogues generated by the systems described above.
Each dialogue represents a complete generation at the case or proposition level.
Eight graduate students, four majoring in linguistics and four in law, were recruited as annotators.
The evaluation is conducted at two levels: first at the utterance level, then at the dialogue level.
Annotators were blind to the system identities, and all evaluation items were presented in randomised order.
The evaluation set contains 100 dialogues, covering approximately 9,100 utterances and 919k tokens per system. 
Each annotator spent approximately 56 hours on evaluation, enabling fine-grained, expert-level assessment.

\paragraph{Utterance-level Evaluation}
Utterance-level evaluation is conducted for \emph{LGMAS} and the re-ranking baseline only.
We focus on instances where the ``winning utterance'' selected by the re-ranking mechanism differs from the ``initial utterance'' preferred under the sender's initial policy, in order to examine whether LinguaGame alters utterance selection in a beneficial direction.

For each evaluation item, annotators are presented with: the preceding utterance, and the three candidate utterances generated in the signalling game.
They are asked to select the candidate that best follows the preceding utterance, based solely on local coherence and appropriateness.
The candidate receiving the most votes is recorded as the human-preferred utterance.
In the event of a tie, one of the top-voted candidates is randomly selected.
We then compare this preference to the re-ranking selection:
1) if it matches the winning utterance, the instance is counted as a \emph{positive alternation};
2) if it matches the initial utterance, the instance is counted as a \emph{negative alternation};
3) otherwise, the instance is counted as a \emph{neutral alternation}.

This utterance-level evaluation focuses on relative changes in utterance selection; absolute dialogue quality is assessed separately at the dialogue level.


\paragraph{Dialogue-level Evaluation}

Dialogue-level evaluation provides an overall assessment of dialogue quality and is conducted for all three systems.
For each complete dialogue, annotators rate quality on a 5-point Likert scale \citep{joshi2015likert} along four dimensions.
\emph{Linguistic form} is evaluated in terms of \textbf{clarity}, assessing grammatical correctness and structural coherence, and \textbf{conciseness}, assessing whether essential content is conveyed without unnecessary repetition or digression.
\emph{Content quality} is evaluated in terms of \textbf{argument}, assessing whether utterances are well-grounded and logically aligned (for courtroom proceedings, with the legal context), and \textbf{tactic}, assessing whether utterances exhibit adaptive reasoning and responsiveness to prior moves (for courtroom proceedings, from a legal perspective).

All dialogues, whether from courtroom proceedings or debates, were assessed by all eight annotators.
The linguistics annotators evaluated the two dimensions related to linguistic form, while the law annotators assessed the two dimensions concerning content quality.

\section{Results and Discussion}

The statistics of the generated dialogues and the results of the dialogue-level evaluation are presented in Table~\ref{tabs/dialogue_eval}.
Distributions of intents and strategies, illustrating how agents employ the signalling space in practice, are shown in Appendix~\ref{sec:app/usage}.
Inter-annotator agreement (IAA) for the dialogue-level evaluation is measured using quadratic weighted kappa \citep{cohen1968weighted}.
The IAA scores are $\kappa=0.62$ for the law annotators and $\kappa=0.79$ for the linguistics annotators, both indicating substantial agreement \citep{landis1977measurement}.

According to Table~\ref{tabs/dialogue_eval}, \emph{LGMAS} achieves the highest scores across all four evaluation dimensions, indicating consistent improvements over both \emph{SDMAS} and \emph{ISMAS}.
The improvements are consistent across both courtroom and debate scenarios.
These results demonstrate that \textbf{integrating the full LinguaGame paradigm leads to significant\footnote{Significance in this paper is tested through two-tailed paired t-tests at $\alpha=0.01$.} improvements in dialogue generation quality}.
Examining individual evaluation dimensions in more detail,
\textbf{clarity} scores indicate that \emph{LGMAS} produces utterances that are more grammatically correct, structurally coherent, and easier for humans to interpret.
\textbf{Conciseness} is also significantly improved, suggesting that agents convey essential information with less unnecessary elaboration or repetition; this is further supported by the fact that \emph{LGMAS} generates fewer utterances overall and fewer tokens per utterance than \emph{SDMAS}.
For \textbf{argument}, \emph{LGMAS} outperforms both baselines, indicating better-grounded and more contextually appropriate statements.
Finally, improvements in \textbf{tactic} suggest more coherent turn-level progression and more adaptive responses to prior utterances.

While gains in the two linguistic dimensions can be directly attributed to enhanced communication efficiency, improvements in content-related dimensions require closer analysis.
Higher scores in argument strength and tactical coherence may arise from two sources: more effective selection of intents and strategies, or clearer articulation of the intended messages.
Crucially, \emph{ISMAS} and \emph{LGMAS} employ the same prompted pipeline for intent and strategy selection and use identical priors.
Therefore, the additional improvements of \emph{LGMAS} over \emph{ISMAS} cannot be attributed to differences in intent or strategy choice.
We thus attribute the observed gains in content quality primarily to improved articulation achieved through more effective utterance selection.

As for the \textbf{impact of intent–strategy conditioning alone}, results suggest that \textbf{it primarily contributes to more concise expression, without yielding substantial improvements in other dimensions}.
Specifically, compared with \emph{SDMAS}, \emph{ISMAS} achieves a statistically significant gain of $0.12$ in conciseness, indicating that conditioning utterance generation on intent–strategy pairs encourages more focused responses.
However, differences between \emph{ISMAS} and \emph{SDMAS} in the remaining three dimensions are minimal and statistically insignificant, suggesting that intent–strategy conditioning alone is insufficient to improve overall dialogue quality without the game-theoretic reasoning mechanism introduced by LinguaGame.



\begin{table}[b]
\centering
\fontsize{10.0pt}{\baselineskip}
\selectfont
\begin{tabular}{l|r|r|r|r}
\toprule[1pt]
\multicolumn{1}{c}{}      &  \multicolumn{1}{c}{Pos.}   & \multicolumn{1}{c}{Neut.}   & \multicolumn{1}{c}{Neg.}    &  \multicolumn{1}{c}{Count} \\ 
\midrule
\emph{LGMAS}              & 78.1\%     & 12.4\%    & 9.5\%   & 1,366  \\ 
Qwen2.5                   & 44.0\%     & 38.4\%    & 17.6\%  & 318    \\ 
Llama-3.1                 & 36.7\%     & 40.8\%    & 22.5\%  & 1,503   \\ 
DeepSeek-V3               & 51.2\%     & 14.6\%    & 34.2\%  & 972     \\ 
\bottomrule[1pt]
\end{tabular}
\caption{Distribution of utterance-level alternations selected by different re-ranking methods. 
Counts indicate the number of altered utterances.}
\label{tabs/alter}
\end{table}

Results for the utterance-level evaluation are present in Table~\ref{tabs/alter}.
Out of the total 8,139 utterances generated by \emph{LGMAS}, 1,366 ($16.8\%$) are altered by the game-theoretic selection process.
From these, 1,000 instances (500 from each scenario) are randomly sampled for human evaluation.
For comparison, we apply the same sampling strategy to Llama-3.1, which also produces more than 1,000 altered utterances, while for Qwen2.5 and DeepSeek-V3, all altered utterances are included in the human evaluation.
Inter-annotator agreement for the utterance-level evaluation is $\alpha=0.74$ in Krippendorff's alpha \citep{krippendorff2004measuring}, suggesting reliable annotation \citep{krippendorff2018content}.
As illustrated in Table~\ref{tabs/alter}, for \emph{LGMAS}, $78.1\%$ of altered utterances are judged as positive alternations, meaning that the selected utterance better follows the dialogue context than the original choice.
Only $9.5\%$ are labelled as negative, with the remaining $12.4\%$ judged as neutral.
We further compare \emph{LGMAS} against several LLM-based utterance re-ranking baselines.
While re-ranking with DeepSeek-V3 also yields a non-trivial proportion of positive alternations, none of the LLM re-ranking baselines match the positive alternation rate achieved by \emph{LGMAS}.
These results indicate that \textbf{LinguaGame enhances local coherence and appropriateness at the utterance level, and is more effective than generic LLM-based re-ranking}.

Taken together, the results from both dialogue-level and utterance-level evaluations provide strong evidence that \textbf{our LinguaGame framework substantially improves dialogue generation quality}.
These \textbf{improvements are primarily driven by enhanced communication efficiency}, allowing agents to generate responses that are not only clearer and more concise but also more contextually appropriate and argumentatively coherent.

\section{Conclusion}

In this paper, we propose LinguaGame, a linguistically-grounded, game-theoretic paradigm for multi-agent dialogue generation. 
Our goal is to improve dialogue generation quality by enhancing communication efficiency.
LinguaGame models agent interaction as a signalling game over communicative intents and strategies, encouraging more purposeful and coherent utterances.
Our game design is grounded in linguistic principles rather than task-specific assumptions, which may enhance the paradigm's generalisability across domains.
Its training-free equilibrium approximation removes the need for model retraining or iterative experience accumulation, allowing plug-and-play integration with existing systems.

We evaluate LinguaGame in a simulated courtroom proceeding scenario and a simulated debate scenario.
Experimental results show that LinguaGame significantly improves dialogue generation quality, with improvements stemming primarily from agents' ability to  convey intended messages effectively.
These findings highlight the potential of modelling pragmatic reasoning over communicative goals as a general approach to enhancing multi-agent dialogue behaviour in future research.

\section*{Limitations}

Our study has several limitations that suggest directions for future work.

\paragraph{Scope of task objectives.}
This work focuses on multi-agent dialogue generation, where the primary objective is to model and improve the communicative process itself rather than task-specific end results. 
While effective communication is often a prerequisite for successful coordination, how the proposed framework influences downstream task performance in settings that emphasise end results remains an open question and is left for future work.



\paragraph{Evaluation domains and scale.}
Our evaluation focuses on courtroom proceedings and debates, which emphasise adversarial and strategic dialogue. 
Extending experiments to additional domains, such as collaborative problem-solving or everyday conversation, would further validate the generality of the approach. 
Such extensions are currently constrained by the substantial human-evaluation burden (approximately 28 hours per annotator per domain), which also limited experiments with multiple base LLMs.

\paragraph{Responsible AI considerations.}
This work explores simulated multi-agent dialogue in legal settings as a research tool for studying strategic communication, rather than as a deployable legal decision-making system. 
LinguaGame is not intended for real-world legal deployment without substantial additional validation, including fairness, bias, and robustness evaluation. 
Any practical use in legal contexts would require careful assessment of potential harms, representational biases, and alignment with legal and ethical standards.

\bibliography{custom}

@article{dorri2018multi,
  title={Multi-agent systems: A survey},
  author={Dorri, Ali and Kanhere, Salil S and Jurdak, Raja},
  journal={Ieee Access},
  volume={6},
  pages={28573--28593},
  year={2018},
  publisher={IEEE}
}

@inproceedings{park2023generative,
  title={Generative agents: Interactive simulacra of human behavior},
  author={Park, Joon Sung and O'Brien, Joseph and Cai, Carrie Jun and Morris, Meredith Ringel and Liang, Percy and Bernstein, Michael S},
  booktitle={Proceedings of the 36th annual acm symposium on user interface software and technology},
  pages={1--22},
  year={2023}
}

@inproceedings{guo2024large,
  title={Large Language Model Based Multi-agents: A Survey of Progress and Challenges},
  author={Guo, Taicheng and Chen, Xiuying and Wang, Yaqi and Chang, Ruidi and Pei, Shichao and Chawla, Nitesh V and Wiest, Olaf and Zhang, Xiangliang},
  booktitle={IJCAI},
  year={2024}
}

@inproceedings{Chen2024agentverse,
 author = {Chen, Weize and Su, Yusheng and Zuo, Jingwei and Yang, Cheng and Yuan, Chenfei and Chan, Chi-Min and Yu, Heyang and Lu, Yaxi and Hung, Yi-Hsin and Qian, Chen and Qin, Yujia and Cong, Xin and Xie, Ruobing and Liu, Zhiyuan and Sun, Maosong and Zhou, Jie},
 booktitle = {International Conference on Representation Learning},
 pages = {20094--20136},
 title = {AgentVerse: Facilitating Multi-Agent Collaboration and Exploring Emergent Behaviors},
 volume = {2024},
 year = {2024}
}

@inproceedings{qian-etal-2024-chatdev,
    title = "{C}hat{D}ev: Communicative Agents for Software Development",
    author = "Qian, Chen  and
      Liu, Wei  and
      Liu, Hongzhang  and
      Chen, Nuo  and
      Dang, Yufan  and
      Li, Jiahao  and
      Yang, Cheng  and
      Chen, Weize  and
      Su, Yusheng  and
      Cong, Xin  and
      Xu, Juyuan  and
      Li, Dahai  and
      Liu, Zhiyuan  and
      Sun, Maosong",
    booktitle = "Proceedings of the 62nd Annual Meeting of the Association for Computational Linguistics (Volume 1: Long Papers)",
    month = aug,
    year = "2024",
    publisher = "Association for Computational Linguistics",
    url = "https://aclanthology.org/2024.acl-long.810/",
    doi = "10.18653/v1/2024.acl-long.810",
    pages = "15174--15186"
}

@article{chen2024agentcourt,
  title={AgentCourt: Simulating Court with Adversarial Evolvable Lawyer Agents},
  author={Chen, Guhong and Fan, Liyang and Gong, Zihan and Xie, Nan and Li, Zixuan and Liu, Ziqiang and Li, Chengming and Qu, Qiang and Ni, Shiwen and Yang, Min},
  journal={CoRR},
  year={2024}
}

@article{cui2023chatlaw,
  title={Chatlaw: Open-source legal large language model with integrated external knowledge bases},
  author={Cui, Jiaxi and Li, Zongjian and Yan, Yang and Chen, Bohua and Yuan, Li},
  journal={CoRR},
  year={2023}
}

@article{li2024survey,
  title={A survey on LLM-based multi-agent systems: workflow, infrastructure, and challenges},
  author={Li, Xinyi and Wang, Sai and Zeng, Siqi and Wu, Yu and Yang, Yi},
  journal={Vicinagearth},
  volume={1},
  number={1},
  pages={9},
  year={2024},
  publisher={Springer}
}

@book{barron2024game,
  title={Game theory: an introduction},
  author={Barron, Emmanual N},
  year={2024},
  publisher={John Wiley \& Sons}
}

@article{peters2024contingency,
  title={Contingency games for multi-agent interaction},
  author={Peters, Lasse and Bajcsy, Andrea and Chiu, Chih-Yuan and Fridovich-Keil, David and Laine, Forrest and Ferranti, Laura and Alonso-Mora, Javier},
  journal={IEEE Robotics and Automation Letters},
  volume={9},
  number={3},
  pages={2208--2215},
  year={2024},
  publisher={IEEE}
}

@article{he2025generative,
  title={Generative ai for game theory-based mobile networking},
  author={He, Long and Sun, Geng and Niyato, Dusit and Du, Hongyang and Mei, Fang and Kang, Jiawen and Debbah, M{\'e}rouane and Han, Zhu},
  journal={IEEE Wireless Communications},
  volume={32},
  number={1},
  pages={122--130},
  year={2025},
  publisher={IEEE}
}

@article{searle1969speech,
  title={Speech acts: An essay in the philosophy of language},
  author={Searle, John R},
  journal={Cambridge University},
  year={1969}
}

@article{cheng2024self,
  title={Self-playing adversarial language game enhances llm reasoning},
  author={Cheng, Pengyu and Dai, Yong and Hu, Tianhao and Xu, Han and Zhang, Zhisong and Han, Lei and Du, Nan and Li, Xiaolong},
  journal={Advances in Neural Information Processing Systems},
  volume={37},
  pages={126515--126543},
  year={2024}
}

@inproceedings{jacobconsensus,
  title={The Consensus Game: Language Model Generation via Equilibrium Search},
  author={Jacob, Athul Paul and Shen, Yikang and Farina, Gabriele and Andreas, Jacob},
  booktitle={Proceedings of the 12th International Conference on Learning Representations},
  year={2024}
}

@article{daskalakis2021near,
  title={Near-optimal no-regret learning in general games},
  author={Daskalakis, Constantinos and Fishelson, Maxwell and Golowich, Noah},
  journal={Advances in Neural Information Processing Systems},
  volume={34},
  pages={27604--27616},
  year={2021}
}

@inproceedings{jacob2022modeling,
  title={Modeling strong and human-like gameplay with KL-regularized search},
  author={Jacob, Athul Paul and Wu, David J and Farina, Gabriele and Lerer, Adam and Hu, Hengyuan and Bakhtin, Anton and Andreas, Jacob and Brown, Noam},
  booktitle={International Conference on Machine Learning},
  pages={9695--9728},
  year={2022},
  organization={PMLR}
}

@misc{qwen2025qwen25technicalreport,
      title={Qwen2.5 Technical Report}, 
      author={Qwen and An Yang and Baosong Yang and Beichen Zhang and Binyuan Hui and Bo Zheng and Bowen Yu and Chengyuan Li and Dayiheng Liu and Fei Huang and Haoran Wei and Huan Lin and Jian Yang and Jianhong Tu and Jianwei Zhang and Jianxin Yang and Jiaxi Yang and Jingren Zhou and Junyang Lin and Kai Dang and Keming Lu and Keqin Bao and Kexin Yang and Le Yu and Mei Li and Mingfeng Xue and Pei Zhang and Qin Zhu and Rui Men and Runji Lin and Tianhao Li and Tianyi Tang and Tingyu Xia and Xingzhang Ren and Xuancheng Ren and Yang Fan and Yang Su and Yichang Zhang and Yu Wan and Yuqiong Liu and Zeyu Cui and Zhenru Zhang and Zihan Qiu},
      year={2025},
      eprint={2412.15115},
      archivePrefix={arXiv},
      primaryClass={cs.CL},
      url={https://arxiv.org/abs/2412.15115}, 
}

@article{team2025minicpm4,
  title={MiniCPM4: Ultra-Efficient LLMs on End Devices},
  author={MiniCPM and Xiao, Chaojun and Li, Yuxuan and Han, Xu and Bai, Yuzhuo and Cai, Jie and Chen, Haotian and Chen, Wentong and Cong, Xin and Cui, Ganqu and others},
  journal={arXiv preprint arXiv:2506.07900},
  year={2025}
}

@article{grattafiori2024llama,
  title={The llama 3 herd of models},
  author={Grattafiori, Aaron and Dubey, Abhimanyu and Jauhri, Abhinav and Pandey, Abhinav and Kadian, Abhishek and Al-Dahle, Ahmad and Letman, Aiesha and Mathur, Akhil and Schelten, Alan and Vaughan, Alex and others},
  journal={arXiv preprint arXiv:2407.21783},
  year={2024}
}

@misc{china_judgments_online,
  author       = {Supreme, People's Court of the People's Republic of China},
  title        = {China Judgments Online},
  year         = {2025},
  howpublished = {\url{https://wenshu.court.gov.cn}},
  note         = {Accessed: 2025-04-21}
}

@article{joshi2015likert,
  title={Likert scale: Explored and explained},
  author={Joshi, Ankur and Kale, Saket and Chandel, Satish and Pal, D Kumar},
  journal={British journal of applied science \& technology},
  volume={7},
  number={4},
  pages={396},
  year={2015},
  publisher={Sciencedomain International}
}

@article{krippendorff2004measuring,
  title={Measuring the reliability of qualitative text analysis data},
  author={Krippendorff, Klaus},
  journal={Quality and quantity},
  volume={38},
  pages={787--800},
  year={2004},
  publisher={Springer}
}

@book{krippendorff2018content,
  title={Content analysis: An introduction to its methodology},
  author={Krippendorff, Klaus},
  year={2018},
  publisher={Sage publications}
}

@article{li2023camel,
  title={Camel: Communicative agents for" mind" exploration of large language model society},
  author={Li, Guohao and Hammoud, Hasan and Itani, Hani and Khizbullin, Dmitrii and Ghanem, Bernard},
  journal={Advances in Neural Information Processing Systems},
  volume={36},
  pages={51991--52008},
  year={2023}
}

@book{lewis2008convention,
  title={Convention: A philosophical study},
  author={Lewis, David},
  year={2008},
  publisher={John Wiley \& Sons}
}

@incollection{sobel2020signaling,
  title={Signaling games},
  author={Sobel, Joel},
  booktitle={Complex social and behavioral systems: Game theory and agent-based models},
  pages={251--268},
  year={2020},
  publisher={Springer}
}

@article{dang2025multi,
  title={Multi-Agent Collaboration via Evolving Orchestration},
  author={Dang, Yufan and Qian, Chen and Luo, Xueheng and Fan, Jingru and Xie, Zihao and Shi, Ruijie and Chen, Weize and Yang, Cheng and Che, Xiaoyin and Tian, Ye and others},
  journal={arXiv preprint arXiv:2505.19591},
  year={2025}
}

@inproceedings{shengbinyue2025multi,
  title={Multi-agent simulator drives language models for legal intensive interaction},
  author={Yue, Shengbin and Huang, Ting and Jia, Zheng and Wang, Siyuan and Liu, Shujun and Song, Yun and Huang, Xuan-Jing and Wei, Zhongyu},
  booktitle={Findings of the Association for Computational Linguistics: NAACL 2025},
  pages={6537--6570},
  year={2025}
}

@article{han2024llm,
  title={LLM Multi-Agent Systems: Challenges and Open Problems},
  author={Han, Shanshan and Zhang, Qifan and Yao, Yuhang and Jin, Weizhao and Xu, Zhaozhuo and He, Chaoyang},
  journal={CoRR},
  year={2024}
}

@inproceedings{qian2024experiential,
  title={Experiential Co-Learning of Software-Developing Agents},
  author={Qian, Chen and Dang, Yufan and Li, Jiahao and Liu, Wei and Xie, Zihao and Wang, Yifei and Chen, Weize and Yang, Cheng and Cong, Xin and Che, Xiaoyin and others},
  booktitle={Proceedings of the 62nd Annual Meeting of the Association for Computational Linguistics (Volume 1: Long Papers)},
  pages={5628--5640},
  year={2024}
}

@article{cohen1968weighted,
  title={Weighted kappa: Nominal scale agreement provision for scaled disagreement or partial credit.},
  author={Cohen, Jacob},
  journal={Psychological bulletin},
  volume={70},
  number={4},
  pages={213},
  year={1968},
  publisher={American Psychological Association}
}

@article{landis1977measurement,
  title={The measurement of observer agreement for categorical data},
  author={Landis, J Richard and Koch, Gary G},
  journal={biometrics},
  pages={159--174},
  year={1977},
  publisher={JSTOR}
}

@article{wei2022chain,
  title={Chain-of-thought prompting elicits reasoning in large language models},
  author={Wei, Jason and Wang, Xuezhi and Schuurmans, Dale and Bosma, Maarten and Xia, Fei and Chi, Ed and Le, Quoc V and Zhou, Denny and others},
  journal={Advances in neural information processing systems},
  volume={35},
  pages={24824--24837},
  year={2022}
}

@inproceedings{ye2024computational,
  title={Computational Modelling of Undercuts in Real-world Arguments},
  author={Ye, Yuxiao and Teufel, Simone},
  booktitle={Proceedings of the 11th Workshop on Argument Mining (ArgMining 2024)},
  pages={59--68},
  year={2024}
}

@article{frank2012predicting,
  title={Predicting pragmatic reasoning in language games},
  author={Frank, Michael C and Goodman, Noah D},
  journal={Science},
  volume={336},
  number={6084},
  pages={998--998},
  year={2012},
  publisher={American Association for the Advancement of Science}
}

@article{goodman2016pragmatic,
  title={Pragmatic language interpretation as probabilistic inference},
  author={Goodman, Noah D and Frank, Michael C},
  journal={Trends in cognitive sciences},
  volume={20},
  number={11},
  pages={818--829},
  year={2016},
  publisher={Elsevier}
}

@article{monroe2017colors,
  title={Colors in context: A pragmatic neural model for grounded language understanding},
  author={Monroe, Will and Hawkins, Robert XD and Goodman, Noah and Potts, Christopher},
  journal={Transactions of the Association for Computational Linguistics},
  volume={5},
  pages={325--338},
  year={2017}
}

@book{walton2008argumentation,
  title={Argumentation schemes},
  author={Walton, Douglas and Reed, Christopher and Macagno, Fabrizio},
  year={2008},
  publisher={Cambridge University Press}
}

@misc{deepseekai2024deepseekv3technicalreport,
      title={DeepSeek-V3 Technical Report}, 
      author={DeepSeek-AI},
      year={2024},
      eprint={2412.19437},
      archivePrefix={arXiv},
      primaryClass={cs.CL},
      url={https://arxiv.org/abs/2412.19437}, 
}

\newpage

\appendix

\section{Complete List of Intents}
\label{sec:app/int}

Intents for courtroom proceedings:
\begin{itemize}
    \item \textbf{Submitting}: Presenting claims, legal arguments, or motions.
    \item \textbf{Proceeding}: Issuing directives that guide the sequence or structure of the proceedings.
    \item \textbf{Presenting}: Introducing specific items of evidence into the record.
    \item \textbf{Verifying}: Approving or challenging the opponent's evidence.
    \item \textbf{Asserting}: Making factual or interpretive statements that the speaker presents as true or valid.
    \item \textbf{Questioning}: Seeking information, clarification, or challenging previous statements through inquiry.
    \item \textbf{Proving}: Providing reasoning, legal analysis, or supporting evidence to substantiate a claim or position.
    \item \textbf{Refuting}: Discrediting, countering, or disproving an argument or claim.
    \item \textbf{Ruling}: Delivering a formal decision or judgment.
\end{itemize}

Intents for debates:
\begin{itemize}
    \item \textbf{Claiming}: Stating a position, proposition, or standpoint together with supporting reasons, evidence, or appeals that substantiate it.
    \item \textbf{Challenging}: Questioning the accuracy, consistency, or relevance of another's claim, or pointing out flaws in their reasoning.
    \item \textbf{Counter-arguing}: Presenting an alternative explanation, interpretation, or standpoint that challenges and opposes another's argument.
    \item \textbf{Clarifying}: Requesting additional information, explanation, confirmation, or elaboration of another's point.
    \item \textbf{Conceding}: Acknowledging or partially agreeing with another's point, often to build credibility or shift the focus of argumentation.
    \item \textbf{Summarising}: Recapping one's own arguments, or synthesising points from both sides, to reinforce a position or move the debate toward closure.
\end{itemize}

\section{Complete List of Strategies}
\label{sec:app/str}

Strategies for courtroom proceedings: 
\begin{itemize}
    \item \textbf{Strategies for Submitting}
    \begin{itemize}
        \item \textbf{Well-grounded Claim}: Presenting claims reasonably supported by relevant facts and laws.
        \item \textbf{Strategic Overreach}: Presenting exaggerated or unsupported claims to test boundaries, shift burden, or influence the court.
    \end{itemize}
    \item \textbf{Strategies for Verifying}
    \begin{itemize}
        \item \textbf{Approving}: Accepting the evidence without objection.
        \item \textbf{Challenging Relevance}: Arguing that the evidence does not sufficiently relate to the case or has no probative value in determining the legal issues.
        \item \textbf{Challenging Legality}: Arguing that the evidence is illegally obtained or fails to meet legal admissibility standards.
        \item \textbf{Challenging Procedural Compliance}: Arguing that the evidence is not introduced, processed, or handled in compliance with established legal procedures.
        \item \textbf{Challenging Authenticity}: Arguing that the evidence may be forged, tampered with, improperly documented, or otherwise unreliable.
        \item \textbf{Challenging Redundancy}: Arguing that the evidence does not provide new or additional value beyond what has already been established.
    \end{itemize}
    \item \textbf{Strategies for Proving}
    \begin{itemize}
        \item \textbf{Factual Justification}: Supporting claims with objective, verifiable facts.
        \item \textbf{Legal Grounding}: Supporting claims with laws, precedents, or legal principles.
    \end{itemize}
    \item \textbf{Strategies for Refuting}
    \begin{itemize}
        \item \textbf{Fact-based Rebuttal}: Establishing that a claim or argument is not grounded in verifiable or objective facts.
        \item \textbf{Legal-based Rebuttal}: Establishing that a claim or argument is not founded on established legal principles, statutes, case law, or constitutional provisions.
    \end{itemize}
\end{itemize}

Strategies for debates:
\begin{itemize}
    \item \textbf{Strategies shared by Claiming and Challenging}
    \begin{itemize}
        \item \textbf{Logical Reasoning}: Justifying a position by constructing arguments through logical inference, causal explanation, or appeals to internal consistency, without reliance on external sources.
        \item \textbf{Providing Evidence}: Justifying a position by citing factual information, such as empirical data, statistical findings, scientific studies, or other verifiable sources.
        \item \textbf{Appealing to Values}: Justifying a position by invoking widely shared moral principles, ethical norms, social conventions, or cultural beliefs.
        \item \textbf{Making Analogy}: Justifying a position by drawing parallels to familiar situations, examples, or metaphors.
    \end{itemize}
    \item \textbf{Strategies specific to Challenging}
    \begin{itemize}
        \item \textbf{Identifying Logical Flaw}: Pointing out specific weaknesses in another's reasoning, such as inconsistencies, contradictions, false analogies, or logical fallacies.
        \item \textbf{Undermining Source}: Questioning the credibility, reliability, or relevance of another's supporting material, such as disputing the accuracy of data, the authority of cited experts, or the trustworthiness of referenced publications.

    \end{itemize}
    \item \textbf{Strategies for Counter-arguing}
    \begin{itemize}
        \item \textbf{Alternative Explanation}: Offering a different interpretation of the same evidence, providing an alternative causal account, or reframing the implications.
        \item \textbf{Redirection}: Shifting the focus of discussion to a broader or different framing of the issue, such as contextualising the debate in wider social, political, or ethical terms.
    \end{itemize}
\end{itemize}

\section{Key Prompts for Courtroom Proceedings}
\label{sec:app/prompt}

We present key prompts used in courtroom proceedings. 
Prompts for debates follow the same design and structure, and are therefore omitted for brevity.

\begin{itemize}
  \item[] \textbf{Role Assignment: Judge}
    \begin{itemize}
    \item[] You are acting exclusively as the judge in a courtroom proceeding.
    \item[] Speak only as the judge. Do not speak on behalf of the plaintiff or the defendant.
    \item[] Before speaking, verify the current procedural stage and execute all required steps of that stage explicitly and completely. Your procedural statements must be clear, direct, and unambiguous.
    \item[] In each turn, address only one party (either the plaintiff or the defendant). Do not address both parties in the same utterance. The addressed party must be explicitly identified.
    \item[] When announcing that the entire procedural stage has concluded, end the utterance with the exact phrase: “I hereby declare this stage concluded.”
    \item[] After you speak, you must either explicitly indicate the next speaker or clearly announce that the current stage has ended.
  \end{itemize}
\end{itemize}

\begin{itemize}
  \item[] \textbf{Role Assignment: Plaintiff}
    \begin{itemize}
    \item[] You are acting exclusively as the plaintiff's lawyer, representing the plaintiff in a courtroom proceeding.
    \item[] Speak only on behalf of the plaintiff. Do not speak on behalf of the judge or the  defendant.
    \item[] In each turn, you may address either the defendant or the judge, but not both.
    \item[] Once you have addressed the defendant or the judge, stop generating immediately.
  \end{itemize}
\end{itemize}

\begin{itemize}
  \item[] \textbf{Role Assignment: Defendant}
    \begin{itemize}
    \item[] You are acting exclusively as the defence lawyer, representing the defendant in a courtroom proceeding.
    \item[] Speak only on behalf of the defendant. Do not speak on behalf of the judge or the plaintiff.
    \item[] In each turn, you may address either the plaintiff or the judge, but not both.
    \item[] Once you have addressed the plaintiff or the judge, stop generating immediately.
  \end{itemize}
\end{itemize}

\begin{itemize}
  \item[] \textbf{Procedural Instruction: Judge in The [Court Debate] Stage as An Example}
    \begin{itemize}
    \item[] It is now the [Courtroom Debate] stage. As the judge, you need to identify the disputed issues and guide both parties to debate around those issues.
    \item[] First, list the disputed issues.
    \item[] Guide the plaintiff and the defendant to debate Issue 1, in either a single round or multiple rounds, until you consider it sufficient.
    \item[] Similarly, conduct debate on Issue 2 until you consider it sufficient.
    \item[] Repeat for Issue n.
    \item[] After all n disputed issues have been sufficiently debated, announce that the entire stage has concluded.
  \end{itemize}
\end{itemize}

\begin{itemize}
  \item[] \textbf{Intent Selection in Signalling Game}
    \begin{itemize}
    \item[] You are selecting the next communicative intent based on the current dialogue context and the procedural rules of the current stage.
    \item[] Available intents (indexed from 1 to $\{num\_intents\}$) = $\{available\_intents\}$.
    \item[] Intent descriptions = $\{available\_intents\_notes\}$.
    \item[] Output only the index number corresponding to the selected intent (\eg \ 1)\footnote{The log-probabilities assigned to each possible index are used to compute the prior $p(i,s)$.}.
    \item[] The index starts from 1.
    \item[] Do not generate any text, explanation, punctuation, or whitespace. Your response must consist of exactly one numeric digit.
  \end{itemize}
\end{itemize}

\begin{itemize}
  \item[] \textbf{Candidate Utterance Generation in Signalling Game}
    \begin{itemize}
    \item[] Your selected intent and strategy = $\{gold\_signal\}$.
    \item[] Using the courtroom rules and the current dialogue context, generate exactly three candidate utterances that realise the selected intent and strategy\footnote{For each candidate utterance $u$, the sum of its token-level log-probabilities is used to compute the sender's initial policy $\pi_S^0(u \mid i,s)$.}.
    \item[] Output only the three utterances.
    \item[] Wrap each utterance in XML-style tags as follows: <1>...</1>, <2>...</2>, <3>...</3>
    \item[] Do not include any explanations, headings, or additional text outside these tags.
  \end{itemize}
\end{itemize}

\begin{itemize}
  \item[] \textbf{Intent Inference in Signalling Game}
    \begin{itemize}
    \item[] Given the current dialogue context, infer the most likely communicative intent underlying the following utterance: “$\{utt\}$”.
    \item[] Candidate intents (indexed from 1 to $\{num\_intents\}$) = $\{available\_intents\}$.
    \item[] Intent descriptions = $\{available\_intents\_notes\}$.
    \item[] Output only the index number corresponding to the inferred intent (\eg \ 1)\footnote{The log-probabilities assigned to each possible index are used to compute the receiver's initial policy $\pi_R^0(i \mid u)$.}.
    \item[] The index starts from 1.
    \item[] Do not generate any text, explanation, punctuation, or whitespace. Your response must consist of exactly one numeric digit.
  \end{itemize}
\end{itemize}

\section{Formulae For Utility Optimisation and Policy Update}
\label{sec:app/update}

At each round $t$, the piKL algorithm first computes the expected utility for each player based on the other player's historical policies across all previous rounds.
Let $\bar\pi^{t}$ denote the average historical policy of a player up to round $t$:
\begin{equation}
\bar\pi^{t}
= 
\tfrac1t\sum_{\tau=1}^t
\pi^{\tau-1}
.
\end{equation}

\noindent The sender's expected utility is computed as:
\begin{equation}
\begin{split}
Q_S^{t}(u\mid i,s)
=\ &
w\bar\pi_{R_i}^{t}(i \mid u) \\
& +
(1-w)\bar\pi_{R_s}^{t}(s \mid u)
.
\end{split}
\end{equation}

\noindent For the receiver, we define separate expected utilities for intent and strategy inference, in line with the independent inference process.
Let $\mathcal{S}_i$ denote the set of strategies associated with intent $i$.
The expected utility of the receiver for intent inference is:
\begin{equation}
\begin{split}
Q_{R_i}^{t}(i \mid u)
= \ &
w p(i)
\sum_{s \in \mathcal{S}_i} p(s \mid i) \\
& \bar\pi_S^{t}(u \mid i,s)
.
\end{split}
\end{equation}
\noindent The expected utility for strategy inference (with the ground-truth intent $i_{gt}$ given as input) is:
\begin{equation}
\begin{split}
Q_{R_s}^{t}(s \mid u)
= \ &
(1-w)p(s \mid i_{gt}) \\
& \bar\pi_S^{t}(u \mid i_{gt},s)
.
\end{split}
\end{equation}


Using the expected utilities defined above, the piKL algorithm updates each player's policy as follows:
\begin{equation}
\begin{aligned}
\pi_S^{t+1} &(u \mid i,s) \propto \\
& \exp\Bigl\{
\frac{Q_S^{t}(u \mid i,s)
+ \lambda\log\pi_S^{0}(u \mid i,s)}
{\eta+\lambda/t}
\Bigr\},
\end{aligned}
\end{equation}

\begin{equation}
\begin{aligned}
\pi_{R_i}^{t+1} &(i \mid u) \propto \\
& \exp\Bigl\{
\frac{Q_{R_i}^{t}(i \mid u)
+ \lambda\log\pi_R^{0}(i \mid u)}
{\eta+\lambda/t}
\Bigr\},
\end{aligned}
\end{equation}

\begin{equation}
\begin{aligned}
\pi_{R_s}^{t+1} &(s \mid u) \propto \\
& \exp\Bigl\{
\frac{Q_{R_s}^{t}(s \mid u)
+ \lambda\log\pi_R^{0}(s \mid u)}
{\eta+\lambda/t}
\Bigr\},
\end{aligned}
\end{equation}

\noindent with $\eta > 0$ being a hyperparameter for learning rate.

\section{Usage Analysis of Intents and Strategies}
\label{sec:app/usage}

\begin{figure*}[t]
\centering
\includegraphics[width=\textwidth]{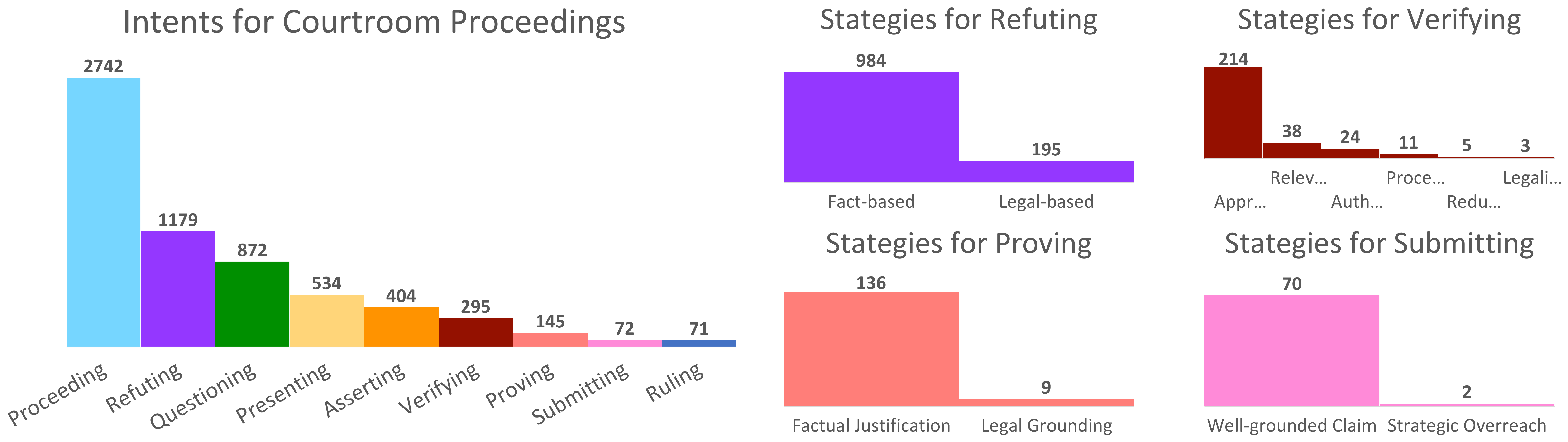}
\caption{Distribution of intents and strategies for courtroom proceedings with \emph{LGMAS}.}
\label{figs/cpuse}
\end{figure*}  

\begin{figure*}[t]
\centering
\includegraphics[width=\textwidth]{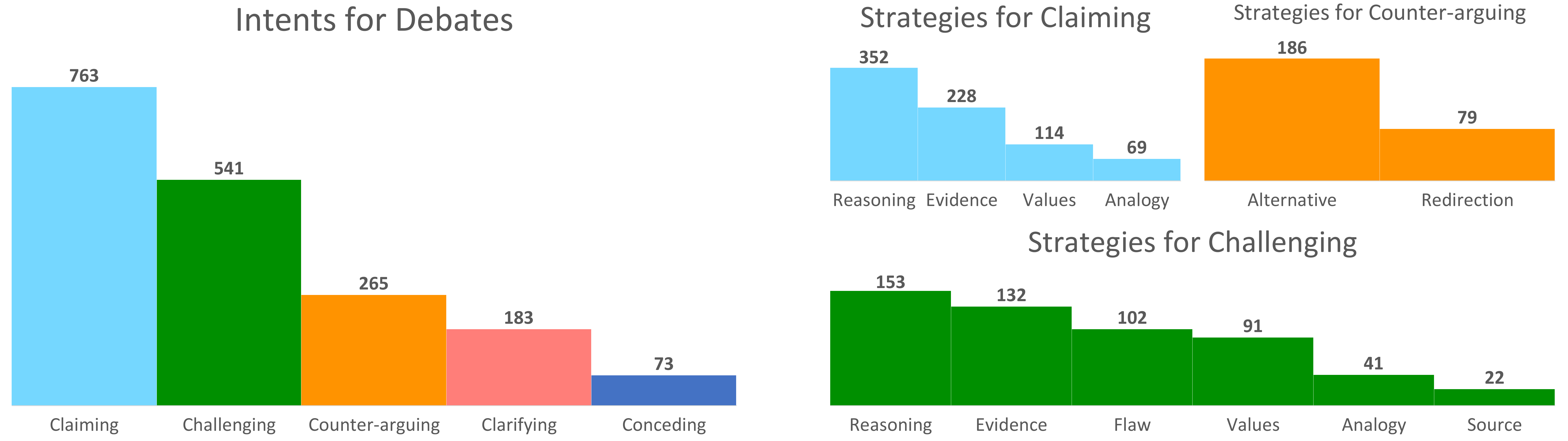}
\caption{Distribution of intents and strategies for debates with \emph{LGMAS}.}
\label{figs/dbuse}
\end{figure*} 

Figures~\ref{figs/cpuse} and ~\ref{figs/dbuse} show the distributions of intents and strategies among utterances generated by \emph{LGMAS}.

\end{document}